\documentclass[sigconf, nonacm]{acmart}

\newcommand\vldbpagestyle{plain} 

\newcommand{\topic}[1]{\noindent \underline{ \bf #1}}

\AtBeginDocument{%
  }

\setcopyright{none} 
\renewcommand\footnotetextcopyrightpermission[1]{}
\begin{document}


\title{A Vision for Semantically Enriched Data Science}

\author{Udayan Khurana$^1$, Kavitha Srinivas$^1$, Sainyam Galhotra$^2$, Horst Samulowitz$^1$}
\affiliation{%
  \institution{$^1$IBM Research, $^2$University of Chicago}
  \country{}
}
\email{ukhurana@us.ibm.com, kavitha.srinivas@ibm.com, sainyam@uchicago.edu, samulowitz@us.ibm.com}





\begin{abstract}
  The recent efforts in automation of machine learning or data science has achieved success in various tasks such as hyper-parameter optimization or model selection. However, key areas such as utilizing domain knowledge and data semantics are areas where we have seen little automation. 
 Data Scientists have long leveraged common sense reasoning and domain knowledge to understand and enrich data for building predictive models.
  In this paper we discuss important shortcomings of current data science and machine learning solutions. We then envision how leveraging {\em semantic} understanding and reasoning on data in combination with novel tools for data science automation can help with consistent and explainable data augmentation and transformation. Additionally, we discuss how semantics can assist data scientists in a new manner by helping with challenges related to trust, bias, and explainability in machine learning. Semantic annotation can also help better explore and organize large data sources.
\end{abstract}


\maketitle
\pagestyle{\vldbpagestyle}
\section{Introduction}

The automation of machine learning (AutoML) or data science processes has received considerable attention recently\footnote{\url{ https://www.tinyurl.com/forbes-automated-data-science/}\hspace{0.2cm}}. 
 The driving force behind AutoML is the desire to reduce human intervention and thereby save time and cost, and improve efficacy of the modeling process. Such automation has recently been achieved through techniques such as reinforcement learning, multi-arm bandits, and numerical optimization methods such as bayesian optimization amongst others. 
While the automation efforts proposed so far have led to efficiencies in different parts of the data science (DS) process, one dimension relatively untouched is that of {\em semantics} in data. Traditionally in data science, where the process is heavily dependent on a human data scientist or a domain expert, semantics of the data plays a crucial role. A data scientist studies the given problem or data and relates them to the concepts in the real world and uncovers relationships between them. This is followed by linking the problem, typically a regression or a classification task defined by the target variable, to the concepts represented by features. Finally, the data scientist utilizes the knowledge of the domain, or general knowledge of the world to perform operations which enhance the modeling capability of the given data, such as feature enhancement. 
Amongst the automation techniques thus far, only few have made progress concerning semantics in data science. Much of this area is still open for further investigation; it is much needed in industry and is of interest in academic research as well. In this paper, we gather the work relevant to semantic data science. We identify opportunities and propose certain ideas for extending automated data science techniques and methods to leverage semantics.

\begin{figure}
    \centering
    \includegraphics[scale=.18]{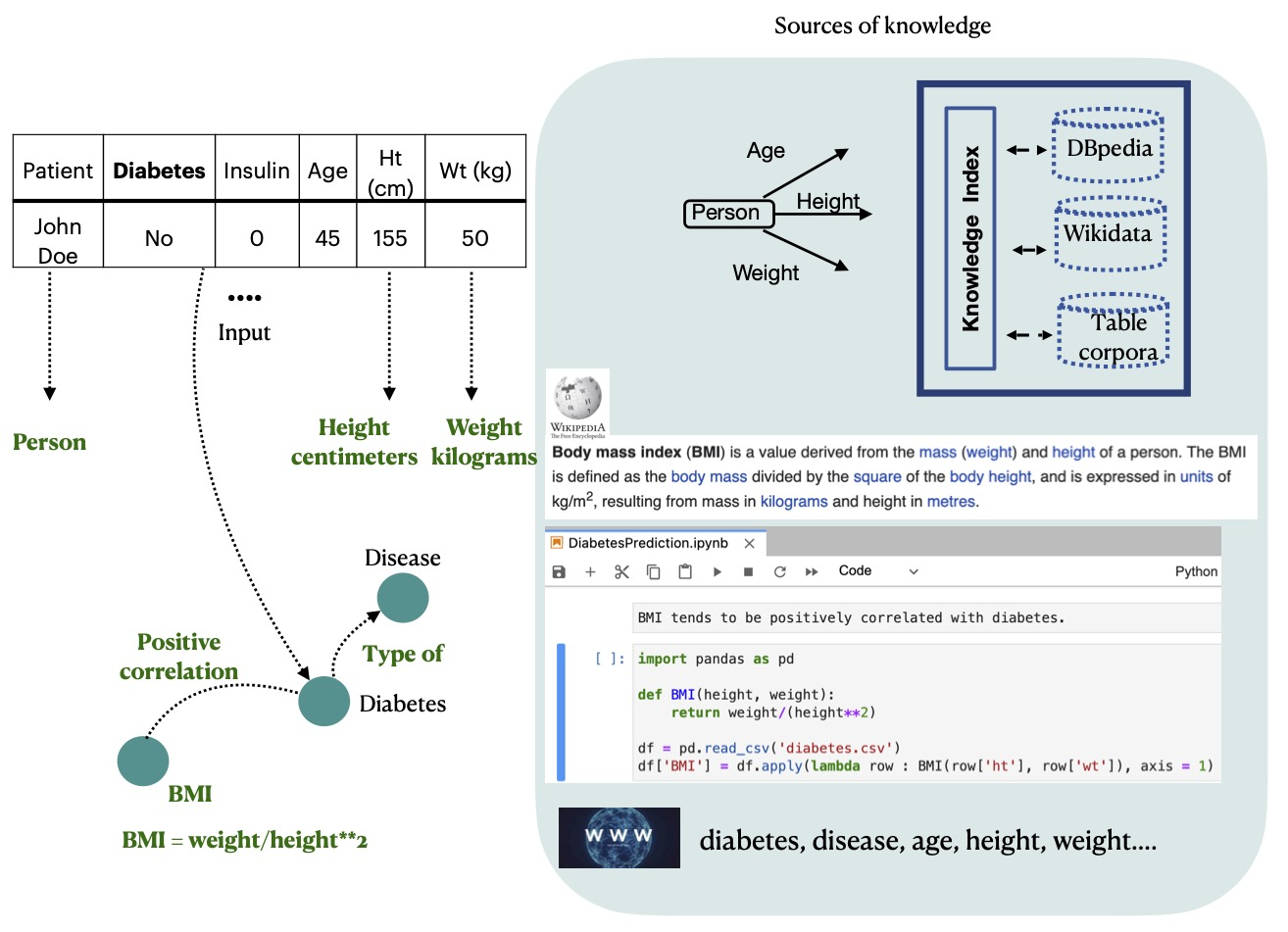}
    \caption{Modeling Diabetes with Semantics}
    \label{fig:example}
\end{figure}

Figure~\ref{fig:example} shows a motivating example for injecting semantics into a data science pipeline.  The input table is about modeling diabetes as a function of various parameters including age, height, weight, etc.  With recent advancements, it is now possible to imbue this data with semantics shown in green labels to help make sense of the data. The right blue panel shows the sources of knowledge available that can be leveraged to add semantics. Knowledge graphs such as DBpedia, Wikipedia, or massive corpora of tables on the web can help make sense of columns. As an example, column to concept mapping technology~\cite{10.1145/3292500.3330993,khurana2020semantic} can help map named entity columns such as ``Patient'' to the concept \texttt{Person}. The Person concept in knowledge graphs frequently has a set of attributes in knowledge graphs, which can be exploited to match ``Ht'' to \texttt{Height}, using approaches for column to property matching.  Units may be approximated by signatures on various attribute distributions in knowledge graphs.  In addition to knowledge graphs, there is a wealth of information buried in unstructured sources such as Wikipedia, text cells of data science notebooks and code. Mining techniques can be used on these corpora to automatically create features for the data science problem at hand.  As an example, the formula for calculating BMI can be mined from both code and text in Wikipedia, and BMI can be used as a semantics-based feature to improve diabetes prediction.  Furthermore, formulae mined from code notebooks can either allow for the extraction of expressions for feature enhancement, or provide hints on when such augmentation is likely to produce a meaningful signal, for instance, if the expression often appears across notebooks working with similar data. Note that these may appear routine for an experienced data scientist, but it is challenging for software to perform these mappings and inferences automatically.  In this paper, we aim to highlight appropriate {\em semantic technology} line of work and project the next important steps.

Data preparation\footnote{sometimes also referred to as {\em feature engineering} or {\em feature enhancement}} is central to any successful model building task. It is the process of altering the feature space representation to better suit model fitting or predictability of the target variable~\cite{fechapter}. It is also a step that heavily involves semantic mapping and reasoning by a data scientist. One example is illustrated in the BMI derivation in Figure~\ref{fig:example}, where we outline how an automation of this process can work via Wikipedia. To show the generalizability of using semantics for data preparation, consider these two other examples drawn from different domains. First, consider loan approval prediction from Kaggle\footnote{\url{https://www.tinyurl.com/yonatanrabinovich-loan}}
where the goal is to predict whether a given loan application will be approved or not. Among the provided attributes are the loan amount, the gender of the applicant, marital status, education, applicant income, co-applicant income, etc. A data scientist knowledgeable in the domain of loans would arrive at ${applicant\_income + coapplicant\_income} \over {loan\_amount}$ as one of the predictors. Performing this step of data preparation requires various levels of semantic operations. This includes primarily mapping given columns to real-world concepts; then, finding relationships between these feature concepts; finally, finding which relationships make sense with respect to the target variable's concept and so on.

Our second example is drawn from Kaggle datasets on COVID-19\footnote{\url{https://www.kaggle.com/imdevskp/corona-virus-report}}.  A common ratio that is used for modeling the disease is the case fatality rate (CFR).  CFR is a formula relating the number of deaths to confirmed cases such that ${\text{CFR}}{\%}={\frac {\text{Number of deaths from disease}}{\text{Number of confirmed cases of disease}}}\times 100$.   Such formulae can be found in both code manipulating COVID notebooks and in Wikipedia, and again can be used to create new features for model building.

In addition to data preparation, other aspects of machine learning (ML) and data science are also dependent on semantic interpretation of data and models. For example, model explainability is based in part on interpretation of concepts in the data and the meaning of operations performed on them. In order to ensure fairness through de-biasing, we require recognition of traditional points of bias~\cite{mehrabi2021survey} and ensuring the same are not repeated in the future. To automate this process, semantic concept recognition and an understanding of the causal dependencies between identified concepts would be central to a trustworthy system that identifies and removes it. Similarly, enforcing business rules through symbolic-AI~\footnote{\url{https://tinyurl.com/inrule-business-rules-ml}}
and semantics enhances the case for semantically-driven automated data science.  In addition, initial work has targeted understanding hyper-parameters and extraction of constraints in applying them \cite{hpodoc}, as well as use of semantics for data cleansing \cite{6982731}.  

In this paper, we first talk about relevant works in semantic understanding and annotation of tabular data (Section~\ref{tabular_data_section}), code (Section~\ref{code_section}) and text (Section~\ref{nlp_expression_section}), as the means of building infrastructure for semantics-oriented data science. 
We then focus our attention on automated data preparation (Section~\ref{autofe_section}) to present popular non-semantic works and the recent developments in semantics-oriented data preparation along with new opportunities. 
This is followed by machine learning and particularly AutoML (Section~\ref{ml_section}).
We conclude with extending these ideas for trustworthy machine learning from a semantic dimension (Section~\ref{trustworthy_section}). 


\section{Understanding Semantics}
\begin{figure}
    \centering   \includegraphics[width=\columnwidth]{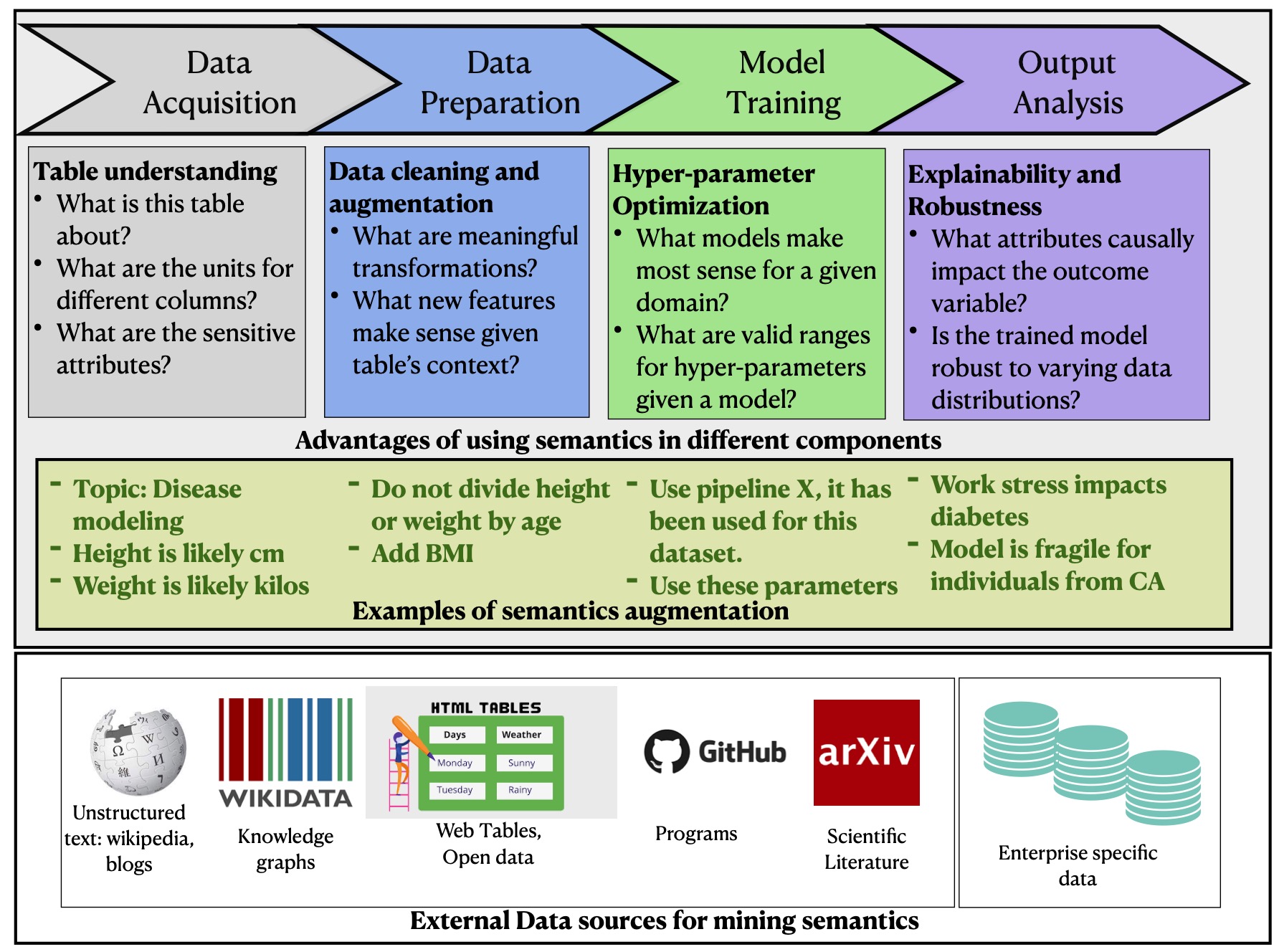}
    \caption{Semantics in a data science pipeline.}
    \label{fig:overview}
\end{figure}
\label{data_section}
The foremost step in data science is the need to understand data. Understanding tables, as shown in Figure~\ref{fig:overview}, in turn helps guide the use of semantics in feature generation, as well as choose models appropriate for the domain.  Explainability in models is enhanced by a knowledge of causal relations between key variables in the domain.  The figure illustrates how semantics can greatly enhance each component in data science workflows. 
Rcent research has demonstrated progess in understanding the semantics of data, which is a crucial first step for semantic data science.
Broadly, the research in this area can be categorized into: (a) table understanding -- which is to map cell values, columns, and indeed entire tables into a well defined set of concepts drawn from ontologies or knowledge graphs; (b) inter-table data discovery -- which discovers for instance, other tables that can be joined or unioned based on a larger index of tables; (c) Understanding semantics in code, text or other forms or less structured data sources. 
 
\subsection{Table Understanding}
\label{tabular_data_section}
Table understanding helps with explainability, trust, bias and feature engineering aspects of data science because it helps pick more suitable features overall to build models. 
 Table understanding can also guide search for relevant data, including tables that make sense to join or union. Here we cover a sample of the literature on table understanding to show that while challenging, it is possible to an extent that is useful for semantic data science.

Table understanding involves identifying the columns type annotation (CTA), identifying the core columns that contain entities of interest, columns that map to properties of these entities (CPA), and mapping tables themselves to broad categories. The goal here is to automatically identify the correct concepts of the columns provided in the data. Additionally, we would like to see the relationships betweeen the columns expressed in the appropriate heirarchy to reflect the intent of the dataset. 
We describe different approaches for table understanding below.

\subsubsection{Knowledge graph based approaches}
Most approaches to table understanding assume the existence of one or more knowledge graphs or ontologies that define the universe of concepts to understand tabular cells or columns.  Columns that contain largely numeric values, dates, social security numbers, etc., are unlikely to appear in knowledge graphs, and are often called L columns.  For columns that contain strings which may potentially map to entities (called NE-columns), tabular understanding is broken into cell entity annotation (CEA), column type annotation (CTA), and column property annotation (CPA), with CEA often being used by many systems to help CTA and CPA.  An additional task considered by many systems is the detection of the subject column (S) which identifies the core entity column, with other columns in the table representing properties of the core entity.  Numerous systems now target these tasks, spurred by the presence of a challenge for these tasks \cite{semtab2019}, with the base knowledge graph being DBpedia or Wikidata.  Example systems that perform one or more of these NE tasks include MantisTable \cite{DBLP:journals/fgcs/CremaschiPRS20}, ColNet \cite{city22932}, TableMiner+ \cite{10.1007/978-3-319-18818-8_25}, and Sherlock \cite{10.1145/3292500.3330993}. Most of these systems are based on neural network based meta-learning on embeddings of column vectors. These approaches are powerful in some cases when plenty of good training examples are available. However, in many cases they do not scale well over the number of column concepts, and particularly on numerical data and in cases where training data is scarce or noisy. $C^2$\cite{khurana2020semantic} presents a maximum likelihood ensemble approach over multiple sources of data such as knowledge bases and noisy webtables and is more scalable in the number of concepts. Some systems, such as Memei \cite{Takeoka_Oyamada_Nakadai_Okadome_2019} jointly train on the CTA task along with column-column relationships to be able to annotate L columns.  

While the approaches and tasks performed vary across systems (e.g., supervised or semi-supervised), in almost all cases, the matching of columns to knowledge graphs is challenging enough for realistic tables. At best, it provides limited support for entirely automated semantic feature engineering.  This is in part due to at least two key limitations of the knowledge based approach: (a) Recall of knowledge bases is severely limited.  In one estimate, only 3\% of the tables contained in the 3.5 billion HTML pages of the Common Crawl Web Corpus can be matched to
DBpedia \cite{10.1145/2872427.2883017}. (b) The approach is especially challenging for L columns, unless the S column detection is perfect and the column heading maps to a knowledge graph property using CPA.  Knowledge agnostic techniques reviewed next circumvent some of the limitations of knowledge graph based approaches for poor recall.

\subsubsection{Knowledge graph agnostic approaches}
Knowledge graph agnostic approaches use large corpora of publicly available tables as knowledge, with schema matching techniques being used to map to tables and columns across the corpus.  The challenge here is how to scale this computation to millions of annotations, to perform what is often termed in the literature as holistic schema matching. Various approaches have been taken to this problem including a binary integration strategy that matches iteratively to create a single vocabulary \cite{10.1145/27633.27634}, or vectorizing column values and using techniques such as locality sensitive hashing (LSH) to minimize the number of pairwise comparisons (\cite{10.1007/978-3-642-35176-1_4}), amongst others.  

More recently, there has been an attempt to vectorize tables from large corpora.  Inspired by work in language models such as BERT~\cite{bert}, which are effective in encapsulating correlations between words in a sentence, some recent works approach the problem of assigning types to columns by building unsupervised BERT based language models for columns based on their values~\cite{trabelsi2020semantic}, adding other intermediate column labels to provide context to improve annotation.  In~\cite{trabelsi2020semantic}, they use 1.6 million tables from the WikiTables domain to build the model.  This approach of using a corpus as a reference knowledge base is clearly attractive because it can greatly alleviate the recall problem of knowledge graphs. The availability of large table corpora such as those gathered from the web\footnote{\url{http://webdatacommons.org/webtables/}}, Viznet\footnote{\url{https://github.com/mitmedialab/viznet}}, Data.gov\footnote{\url{https://www.data.gov/}} make this approach promising and feasible.


\subsection{Inter-table discovery}
\label{subsec:intertable}
While features can often be constructed based on a single table, it is more common in data science that the scientist joins or unions data across tables to compute a model.  Table similarity detection targets this space of data wrangling.  Table similarity detection based on a number of metrics such as the number of matching rows, number of matching columns, the amount of information gained by possibly combining related tables, etc., are used by systems such as Juneau to increase the number of possible features for feature engineering \cite{10.1145/3318464.3389726}.  The problem of finding interesting joins in a data lake is addressed by JOSIE~\cite{10.1145/3299869.3300065}, which performs an efficient top-k overlapping set similarity search over millions of tables.  Joinable tables are ranked by the size of overlap to a queried key column.  JOSIE works on equi-joins.  PEXESO \cite{Dong2021EfficientJT} performs joins on columns with string values using word embeddings, because joinable columns often have slight mismatches in the string representation.  PEXESO is also evaluated for data enrichment; the system shows 1.9\% higher micro-F1 score and 10\% lower mean squared error on machine learning tasks. KAFE~\cite{galhotra2019automated} and S3D~\cite{s3d} create effective joins and unions with knowledge graphs and tables large data lakes by identifying semantic concepts of the given data with those of the database table columns.

\subsection{Mining Semantics in Code and Text}
Assuming we have the technologies to find useful sources of data to model a target variable, automated feature engineering still requires the mining of formulas that allow us to construct meaningful features for a given target, as illustrated in the BMI, CFR or loan examples.  We describe two areas for such mining - one from code, and another from text.

\subsubsection{Data science through mining code:}
\label{code_section}
Recent work has targeted mining code repositories to help the data scientist.  Wranglesearch\footnote{\url{https://www.josecambronero.com/pdf/wranglesearch-draft.pdf}} is a system that mines Kaggle notebooks using dynamic analysis to extract functions and expressions that may help with data wrangling, and is perhaps the most relevant work for automated feature engineering.  Others target the larger data science development space.  \cite{10.1145/3447548.3467455} for instance mines Kaggle notebooks and scripts to find what they call decision points, where alternative models or parameters might be considered, and provide the data science users with these alternatives, so that they might build more robust models.  SOAR \cite{9402016} uses embedding techniques on the documentation of API calls combined with programming by example type approaches to suggest refactoring of data science code from one API to another (e.g. from tensorflow to PyTorch).

\subsubsection{Mining expressions from text:}
\label{nlp_expression_section}
Vast amounts of domain knowledge useful for data science is buried in text as well.  There has been work that targets the extraction of topics for mathematical equations \cite{Yasunaga_Lafferty_2019} by building joint neural models for the textual context around the equation along with the equation itself.  In a data science context, it would be more useful for searching relevant expressions to surface to a data scientist, or to extract features if the terms of the equation can be mapped to columns successfully.  \cite{10.1145/3366423.3380218} describe the use of modified information retrieval techniques on an extracted database of several million equations; they, for instance, find equations associated with {\em Jacobi Polynomial}, or complete equations.  Once again, from a data science perspective, \cite{10.1145/3366423.3380218} connect equations to textual topics, but its utility in an entirely automated feature engineering mechanism is still an open question.

A related area of work is with respect to mining causal relationships from text; causality often determines which features may be semantically meaningful to model a target variable.  Work related to automated causal extraction from text is rich (e.g., \cite{LI2019512}, \cite{hassanzadeh2019answering}). Extraction of such information from notebooks and domain specific corpora could help with feature selection.

\subsection{Utilizing Semantics from Large Language Models}
Inspired by the success of transformer based models in natural language processing (NLP), there are now a number of different works targeting language models for tabular data (see \cite{ijcai2022p761} for a review).  These language models target a number of downstream tasks varying from summarizing the table in text, to column property annotation, SQL generation or entity linking.  While the initial results in this space are interesting, there are numerous open questions in this area: (a) there is a dearth of a large set of downstream tasks that has powered signficant progress in NLP; recent NLP models train on many hundred tasks (e.g., BigBench \footnote{\url{https://github.com/google/BIG-bench}}.  No such standard suite exists for tabular tasks, and each work targets a different downstream task. (b) existing public corpora tend to be web tables which tend to be quite different in their characteristics than enterprise databases; spreadsheets when used are not publicly available to the community, (c) sequence length restrictions of transformer models seem especially challenging for tabular data consisting of millions of rows/columns, (d) numerical columns are frequently treated as text, or specialized (e.g., representing magnitude, first digit and last digit) that may not capture column characteristics properly. 

\section{Data Preparation}
\label{autofe_section}


Data Preparation component of a data science pipeline aims to integrate data acquired from numerous sources, perform data cleaning and formatting to homogenize the available datasets. Data preparation often consists of a multitude of components.
We give an overview of these components in Section~\ref{autofe_nonsem} and discuss their impact on the downstream analytics.
We then describe preliminary work towards using semantics and present how there is a missed opportunity in incorporating semantics. Next, we present concrete ideas to extend various semantic-oblivious data preparation approaches with semantics.
\subsection{Data Preparation Tasks}\label{autofe_nonsem}
Data preparation generally comprises many different steps to integrate, clean, transform and validate the available datasets. We present three representative categories of data preparation tasks.

\noindent \textbf{Data discovery and integration:} These components aim to leverage external data sources to identify new datasets that can be used for downstream analysis. The key challenge of these components is to search useful data over millions of available options and integrating information collated from heterogeneous sources. Prior data discovery and integration techniques have not considered the use of semantics to identify useful datasets. These techniques often generate many false positive, i.e. datasets that are either irrelevant or semantically incoherent.

\noindent \textbf{Data validation and transformation: }
Data transformation techniques either modify and homogenize data values using transformation functions. For instance, they may change the unit of weight from kgs to lbs or generate new attributes by combining the attributes already present such as the addition of BMI using height and weight attributes. Many diverse approaches have focussed towards generation of new attributes.
FICUS~\cite{MarkovitchRosenstein02} performs a beam search over the space of possible features, constructing new features by applying constructor functions. ExploreKit~\cite{DBLP:conf/icdm/KatzSS16} expands the feature space explicitly, one feature at a time. AlphaD3M~\cite{alphad3m} and Cognito~\cite{khurana2016cognito} use reinforcement learning to optimize feature transformation through an agent. LFE~\cite{lfe} directly predicts the most useful transformation per feature based on learning effectiveness of transforms on sketched representations of historical data through a perceptron. 
More details on such approaches can be found in these  surveys~\cite{fechapter},~\cite{zoller2021benchmark} and~\cite{sondhi:2009}.

The following works consider semantics to some extent for data enhancement. Friedman et al.~\cite{friedman2018recursive} and Galhotra et al.~\cite{galhotra2019automated} recursively generate new features using a knowledge base for text classification and web tables, respectively. FeGeLOD~\cite{PH} extracts features from Linked Open Data via queries. These approaches, however, fail to thoroughly explore the available information from multiple data sources. S3D~\cite{s3d} is able to bring in data from multiple sources such as webtables and multiple knowledge graphs but adds new columns irrespective of the target variable. Semantic Feature Discovery~\cite{sfd} uses formulae from a historical repository of data science code files and notebooks by linking concepts in a given data problem with those in the historical code. 
While these approaches are promising, their impact is bounded due to the limited availability and richness of knowledge bases, and the quality of underlying concept matching algorithms. Also, it is an open problem on how to make connections between data and expressions considering different backgrounds and domains. 

\noindent \textbf{Data filtering and cleaning} techniques aim to remove noisy values, fill missing values and replace inaccurate records. Prior techniques for data cleaning explore different data quality issues like presence of duplicates, violation of functional dependencies, outlier detection, etc. Majority of these techniques do not explore the use of semantics and information available from external data sources to improve these components.

\subsubsection{Missed Opportunities due to lack of Semantics in Data Preparation}
The state-of-the-art feature engineering systems mentioned above rely on trial and error for generating new features through transformation functions on the provided features. They try to maximize accuracy but are oblivious of the actual semantics of the data and transformation functions. This has several drawbacks. 
Firstly, they cannot utilize external knowledge to perform specific transformations on data which might be very difficult to stumble upon using generalized exploration techniques. For example, BMI (Figure~\ref{fig:example}) is a complex formula which any of these techniques will take a very long time and number of trials to discover. Secondly, they could violate physical laws such as addition of {\texttt height} and {\texttt weight}, which would be a valid exploration option in these methods but through worldly knowledge, we know that it does not physically make sense and should be avoided. Third, these methods do not have the capability of bringing additional data not provided in the problem but available externally and help improve the modeling capacity. 
Existence of abundance of knowledge in the form of knowledge bases, digital books, wikipedia, blogs, and data science code and so on is a missed opportunity for automated data preparation or feature engineering.

\subsection{Proposal for Utilizing Semantics in AutoFE}
Cherrier et al.~\cite{cherrier2019consistent} use a genetic programming formulation for feature construction, but with constraints in the domain of physics. Examples of constraints are -- energy and dimension can not be added as it does not make sense physically, and performing square after square root does not make much sense either, and so on. We argue a generalization of this approach on top of existing non-semantic approaches can significantly benefit latter's efficacy. Whether it is feature engineering through genetic programming as advised by~\cite{cherrier2019consistent} or through other optimization methods described in Section~\ref{autofe_nonsem} such as Cognito~\cite{khurana2018feature}, DSM~\cite{kanter:dsm}, ExploreKit~\cite{DBLP:conf/icdm/KatzSS16} or even metalearning approach such as LFE~\cite{lfe}, by identifying semantic types and enforcing certain constraints (some of which are global, whereas some specific to the problem) that can be expressed in an appropriate grammar are beneficial. Any of these methods can be restricted in their (non-beneficial) operations by simply checking for semantics or constraint consistency. 
This would require a common grammar to represent such constraints and an effective engine to enforce them. It might of interest for researchers in this area to develop a language that is expressive enough to capture constraints from various domains but simple enough to specify them easily or translate existing business or scientific rules to the new language.

\section{Machine Learning with Semantics}
\label{ml_section}
Much of current ML (including Deep Learning) is operationally devoid of semantics. There is no explicit notion (e.g., as part of the input) of what a feature, value, state, object or label actually `means'. As a result, semantics are not utilized explicitly to build, tune or deploy models.
This includes even the powerful foundational models for language (e.g., GPT3~\cite{gpt3}), speech, images (e.g., Imagenet~\cite{imagenet}), or video recognition, that display impressive performance on specific tasks.
While such models are indeed able to generate effective responses which might seem semantically aware on occasions, the underlying models have virtually no intrinsic understanding of semantics. These ML models are neither provided with, nor infer semantics of the input data. The models are mostly based on statistical and mathematical properties of data. While a data-driven way of deriving insights is desired, the meaning of data itself beyond statistical understanding has been neglected and the resulting issues are well known ranging from bias and explainability~\footnote{\url{https://venturebeat.com/datadecisionmakers/bias-in-ai-is-spreading-and-its-time-to-fix-the-problem/}}, spurious correlations~\footnote{\url{http://www.tylervigen.com/spurious-correlations}}, inappropriate/meaningless responses~\footnote{\url{https://medium.datadriveninvestor.com/chatbot-conversations-gone-wrong-78f7c1af4401}}. 

It is challenging to incorporate semantics because common-sense reasoning has not yet been accomplished in AI. However, we argue that even limited amounts of semantics can help build better AI models. For instance, in otherwise model-free reinforcement learning~\cite{Sutton1998}, if a model knows beforehand that a state-action pair is not valid in reality, then there is no need to explore this scenario. This could avoid unnecesary computation as well as possible model suboptimality. 
Similarly, if it is known that a semantic relationship does not exist between features, one could consider not connecting them in a neural network to possibly achieve better model explanations. As illustrated in Figure~\ref{fig:example}, semantically driven feature engineering based on data transformations are tansparent to the data scientist, regulators, auditors and the general public. For example, there is social, legal and ethical value in understanding why a computational model refuses or accepts a loan application. 

Note that various problem areas in AI actually require much more explicit definitions of the task as input. For instance, AI planning and constraint programming are based on user supplied definitions (e.g., variables, constraints, and objectives). In those settings, at least a part of the overall data semantics is provided as input. However, it is well-known that modeling problems in such ways put a substantial burden on users, since the problem formulation is often not straightforward at all (e.g.,~\cite{mipformulations, modelingCP}). Hence a general semantic automation will benefit these areas as well. 

\subsection{AutoML}
The impact of semantics can also be leveraged as part of the AutoML optimization itself. 
Hyperparameter search is one aspect of AutoML that has been targeted for the use of semantics.  The search space for finding the correct model is large, and the focus here is on using existing artifacts around machine learning to perform a more targeted search through the space of hyperparameters for a model.  \cite{10.1145/3447548.3467455} for instance, mines Kaggle notebooks and scripts to find what they call decision points, where alternative models or parameters might be considered, and provide the data science users with these alternatives, so that they might build more robust models.  
Another approach examines if we can mine data science code for better pipeline search for unseen datasets~\cite{KGPIP}.

We further believe that within hyper-parameter optimization, one could analyze code and Python DocStrings to not only determine available hyper-parameter settings automatically~\cite{hpodoc}, but also to extract constraints on hyper-parameter configurations such as mutually exclusive settings. This at least reduces the search space to valid choices automatically. 
At the pipeline configuration level, one can observe patterns in existing code such as PCA being nearly always preceeded by normalization and not the other way around and use them for instance as a way to prioritize specific pipelines in the automated pipeline exploration.

\begin{figure*}[ht]
    \centering
    \includegraphics[trim={0 0cm 0 0cm},clip, scale=.2]{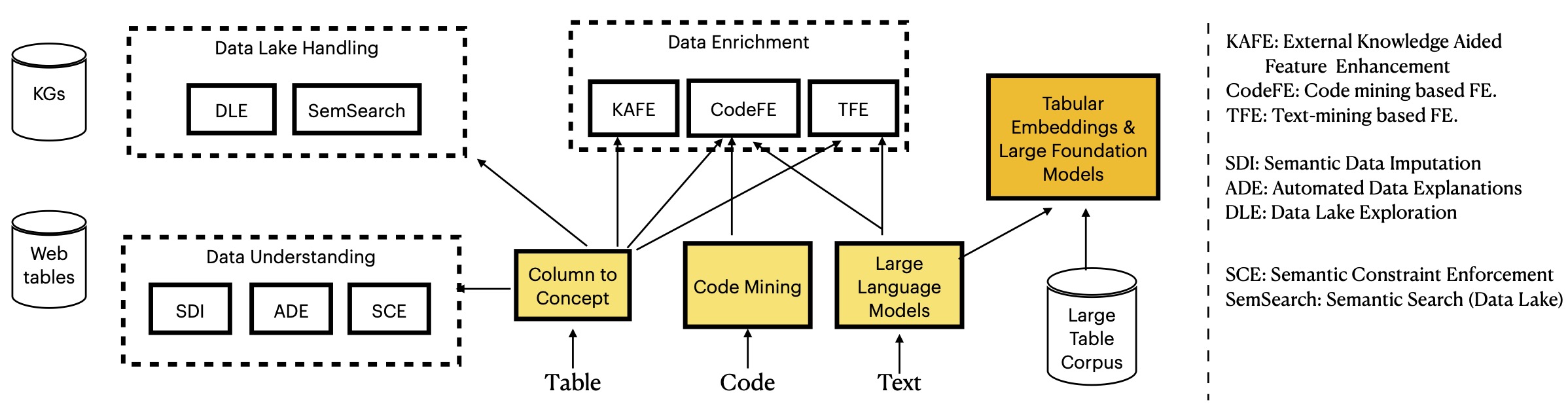}
    \caption{Proposed Components of Semantics Machinery for Data Science. Light yellow filled blocks are the core semantic building blocks, while the dark yellow is a potential core building block (Foundation Tabular Model).}
    \label{fig:arch}
\end{figure*}

\section{Using Semantics for Trustworthy Output Analysis }
\label{trustworthy_section}
The use of machine learning techniques to make high-stakes decisions has raised the importance of trust, which is generally captured in terms of fairness, explainability, robustness, among others. Most of the prior techniques study societal impact of Machine learning by exploring correlation between dataset attributes without understanding their true meaning. 
Consequently, fairness-aware learning techniques are brittle to noise in datasets and do not generalize to real-world settings where datasets may suffer from selection bias, and may even contain spurious correlations. Explainable AI techniques generate explanations that are not semantically coherent~\cite{ustun2019actionable} and are difficult to comprehend. 

\subsection{Causal Dependencies to Infer Semantics}
One of the initial steps towards enabling semantically coherent learning has been to leverage  causal dependencies between attributes. 
These studies have explored the use of causal information in two different ways. (i) Semantic constraints: These techniques impose constraints over input attributes to ensure that the studied combinations of attribute values are feasible~\cite{mothilal2020explaining}. For example, it is not possible that a record with Age $<18$ has Marital Status = Married.
(ii) Dependency information: These techniques evaluate the dependencies between attributes as external knowledge~\cite{karimi2020model,galhotra2021explaining}. For example, increasing salary generally means an increase in savings.

Causal relationships do not capture the meaning of a particular feature but rather the dependencies between them. Consideration of causal dependencies is a step towards leveraging semantics for analytics. 
These techniques show significant promise to incorporate causal information in ML but also raise numerous challenges. 

\noindent \textbf{How to identify attribute dependencies?} Causal dependencies between attributes are estimated from observational data, and it is often inaccurate due to the limited size of available data. Domain scientists need to manually validate semantic constraints/ causal dependencies as a post-hoc analysis to make sure that the analysis is causally consistent.

\noindent \textbf{How to incorporate semantic constraints and dependencies?}
Causation requires testing the effects of changes in attributes, which is often referred to as treatment or intervention. Incorporating interventions into the traditional correlation-based frameworks requires development of new techniques. One of the recent techniques has studied the use of causal notions of fairness and explored estimation of causal effects without any external information.

\noindent \textbf{How can external information help?}
Most techniques rely on domain scientists to either impose causal constraints or input dependency information in the form of a causal graph. Reducing the dependence of such techniques on human players to provide external information is an important challenge that needs more work. External sources of information like data lakes, Knowledge Graphs contain useful information that can be helpful to infer dependencies. 

\section{Open Research Problems}
Of the various research opportunities in semantic data science discussed in this paper, we summarize some of the most promising directions for research.

\topic{RP1:} Inferring semantics from numeric data: (a) Find suitable embeddings to encode numeric columns which capture the concepts, units and quantity referred to by columns containing numerical values. Investigate the need for techniques that use sketching ideas along with summarization methods.
(b) Using numeric embeddings to map columns to their concepts, as it is often performed for textual data.
(c) Create realistic human annotated benchmarks for column to concept mapping, especially for numeric data.
 
\topic{RP2:} Develop techniques to understand and organize large volumes of data (data lakes) through semantic alignment and classification. This includes semantic search and data browsing to help users make sense of very large volumes of schema-on-read, unorganized data. (DLE and SemSearch in Figure~\ref{fig:arch})

\topic{RP3:} Mine large repositories of code for data transformation and attach concepts to the variables of those transformations. 

\topic{RP4:} Mine large volumes of expert text repositories such a Wikipedia, e-books, blogs etc., for actionable information that can be tied to concepts represented in knowledge graphs.

\topic{RP5:} Improve data imputation techniques by concept identification and historical value lookup for those concepts. (SDI in Figure~\ref{fig:arch}).

\topic{RP6:} Discover concepts that reflect bias in model construction depending on the context of the data and task specification. Adjust model training to eliminate bias.



\bibliographystyle{ACM-Reference-Format}
\bibliography{main}
\end{document}